\newif\ifcomments
\commentstrue

\documentclass{article}
\PassOptionsToPackage{square, numbers, compress,sort}{natbib}
\usepackage[final]{nips_2017}

\usepackage[T1]{fontenc}    
\usepackage{booktabs}       
\usepackage{amsfonts}       
\usepackage{nicefrac}       
\usepackage{microtype}      
\usepackage[utf8]{inputenc}
\usepackage{color}
\usepackage{graphicx}
\usepackage{algorithm,algorithmic}
\usepackage{hyperref}
\usepackage{url}            
\usepackage{amsfonts}
\usepackage{amsmath}
\usepackage{booktabs}
\usepackage{wrapfig}

\renewcommand{\paragraph}[1]{

\textbf{#1}}

\usepackage{caption}
\DeclareCaptionType{noticebox}
\usepackage{subcaption}

\newcommand{\titl}{Memory-Efficient Implementation of DenseNets}
\newcommand{\authorinfo}{
  Geoff Pleiss\thanks{Authors contributed equally.}\\
  Cornell University\\
  \texttt{geoff@cs.cornell.edu}
  \And
  Danlu Chen\normalfont\textsuperscript{*}\\
  Fudan University\\
  \texttt{dc688@cornell.edu}
  \And
  Gao Huang, Tongcheng Li\\
  Cornell University\\
  \texttt{\{gh349,tl486\}@cornell.edu}
  \And
  Laurens van der Maaten\\
  Facebook AI Research\\
  \texttt{lvdmaaten@fb.com}
  \And
  Kilian Q. Weinberger\\
  Cornell University\\
  \texttt{kwq4@cornell.edu}
}

\title{\titl}
\author{\authorinfo}

\ifcomments\newcommand{\comments}[1]{#1}\else\newcommand{\comments}[1]{}\fi
\definecolor{clrgp}{rgb}{.9,0,.9}

\begin{document}

\maketitle

\begin{abstract}
  The DenseNet architecture \cite{huang2017densely} is highly computationally efficient as a result of feature reuse.
  However, a na\"ive DenseNet implementation can require a significant amount of GPU memory: If not properly managed, 
  pre-activation batch normalization \cite{he2016identity} and contiguous convolution operations can produce feature maps that grow quadratically with network depth. 
  In this technical report, we introduce strategies to reduce the memory consumption of DenseNets during training.
  By strategically using shared memory allocations, we reduce the memory cost for storing feature maps from \emph{quadratic} to \emph{linear}.
  Without the GPU memory bottleneck, it is now possible to train extremely deep DenseNets.
  Networks with $14M$ parameters can be trained on a single GPU, up from $4M$.
  A $264$-layer DenseNet ($73M$ parameters), which previously would have been infeasible to train, can now be trained on a single workstation with  8 NVIDIA Tesla M40 GPUs.
  On the ImageNet ILSVRC classification dataset, this large DenseNet obtains a state-of-the-art single-crop top-1 error of $20.26\%$. 
\end{abstract}

\section{Introduction}

The DenseNet architecture \cite{huang2017densely} is highly efficient, both in terms of parameter use and computation time.
On the ImageNet ILSVRC-2012 dataset \cite{russakovsky2015imagenet}, a 201-layer DenseNet achieves roughly the same top-1 classification error as a 101-layer Residual Network (ResNet) \cite{he2016deep}, while using half as many parameters ($20M$ vs $44M$) and half as many floating point operations ($80B$/image vs $155B$/image).
Each DenseNet layer is explicitly connected to all previous layers within a pooling region, rather than only receiving information from the most recent layer.
These connections promote feature reuse, as early-layer features can be utilized by all other layers.
Because features accumulate, the final classification layer has access to a large and diverse feature representation.

This inherent efficiency makes the DenseNet architecture a prime candidate for very-high capacity networks.
\citet{huang2017densely} report that a $161$-layer DenseNet (with $k=48$ features per layer and $29M$ parameters) achieves a top-1 single-crop error of $22.2\%$ on the ImageNet ILSVRC classification dataset.
It is reasonable to expect that larger networks would perform even better.
However, with most existing DenseNet implementations, model size is currently limited by GPU memory.

Each layer only produces $k$ feature maps (where $k$ is small -- typically between $12$ and $48$), but uses all previous feature maps as input. This causes the number of \emph{parameters}  to grow quadratically with network depth. It is important to note that this quadratic dependency of parameters w.r.t. depth is \emph{not} an issue, as networks with more parameters perform better and in that respect DenseNets are more competitive than alternative architectures, such as e.g. ResNets. Most na\"ive implementations of DenseNets do however also have a quadratic  memory dependency with respect to \emph{feature maps}. This growth is responsible for the vast majority of the memory consumption, and as we argue in this report, it is implementation issue and not an inherent aspect of the DenseNet architecture. 

The quadratic memory dependency w.r.t. feature maps originates from  intermediate feature maps generated in each layer, which are the outputs of batch normalization and concatenation operations. Intermediate feature maps are utilized both during the forward pass (to compute the next features) and during back-propagation (to compute gradients).  If they are not properly managed, and are na\"ively stored in memory,  training large models can be expensive -- if not infeasible.


In this report, we introduce a strategy to substantially reduce the training-time memory cost of DenseNet implementations, with a minor reduction in speed.
Our primary observation is that \emph{the intermediate feature maps responsible for most of the memory consumption are relatively cheap to compute}.
This allows us to introduce Shared Memory Allocations, which are used by all layers to store intermediate results. Subsequent layers overwrite the intermediate results of previous layers, but their values can be re-populated during the backward pass at minimal cost. 
Doing so reduces feature map memory consumption \emph{from quadratic to linear}, while only adding $15-20\%$ additional training time.
This memory savings makes it possible to train \emph{extremely large} DenseNets on a reasonable GPU budget.
In particular, we are able to extend the largest DenseNet from 161 layers ($k=48$ features per layer, $20M$ parameters) to 264 layers ($k=48$, $73M$ parameters).
On ImageNet, this model achieves a single-crop top-1 error of $20.26\%$,
which (to the best of our knowledge) is state-of-the-art.

Our memory-efficient strategy is relatively straightforward to implement in existing deep learning frameworks.
We offer implementations in Torch\footnote{
  Torch implementation: \url{https://github.com/liuzhuang13/DenseNet/tree/master/models}
} \cite{torch}, PyTorch\footnote{
  PyTorch implementation: \url{https://github.com/gpleiss/efficient_densenet_pytorch}
} \cite{pytorch}, MxNet\footnote{
  MxNet implementation: \url{https://github.com/taineleau/efficient_densenet_mxnet}
} \cite{mxnet}, 
and Caffe\footnote{
  Caffe implementation: \url{https://github.com/Tongcheng/DN_CaffeScript}
} \cite{caffe}.

\section{The DenseNet Architecture}

\begin{figure}
  \centering
  \includegraphics[width=0.8\linewidth]{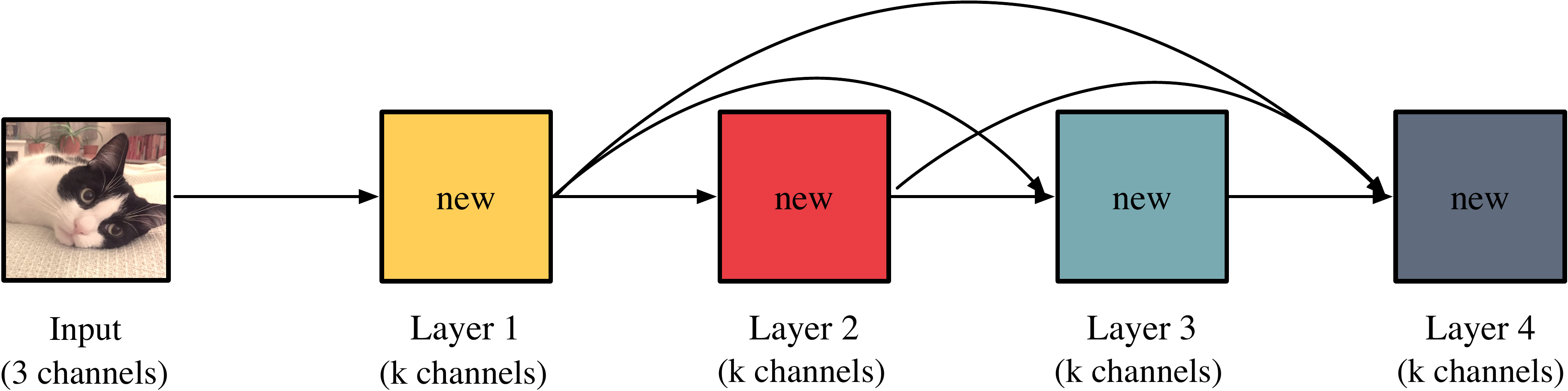}
  \caption{High-level illustration of the DenseNet architecture.}
  \label{fig:basic_densenet}
\end{figure}

\newcommand{\xb}{\mathbf{x}}

At a high-level, a DenseNet explicitly connects all layers with matching feature size. Huang et al. refer to a block of directly connected layers as a \emph{dense block}, which is typically followed by a pooling layer (which reduces feature map sizes) or the final classifier. 
Layers in traditional neural networks only use the most recent features; in a DenseNet, a layer has access to all preceding features within its dense block. Mathematically, $\xb_\ell$ -- the features produced by layer $\ell$ -- are computed as
$$ \xb_\ell = H_\ell \left( [\xb_1, \ldots, \xb_{\ell-1}] \right), $$
where $H_\ell$ denotes the operations of layer $\ell$ and $[\cdot]$ represents the concatenation operator.
\citet{huang2017densely} define $H_\ell$ to be a composite of three operations: batch normalization \cite{ioffe2015batch}, a rectified linear unit (ReLU), and convolution -- in that order.
($H_\ell$ may also include additional operations, such as a 
``bottleneck''; see \cite{huang2017densely}.)
In most deep learning frameworks, each of these three operations produce intermediate feature maps.

Given a dense block with $m$ layers, the input to its final layer is $[\xb_1, \ldots, \xb_{m-1}]$ -- all previous convolutional features.
Because the number of convolutional features grows linearly with network depth, storing these in memory does not impose a significant memory burden. 
However, if the network stores the intermediate feature maps as well (e.g. the batch normalization output), GPU memory may become a limited resource. This is due to the fact that the intermediate features are computed for each input feature map, thus incurring $O(m^2)$ memory usage \emph{if they are stored}. 
Many deep learning frameworks keep these intermediate feature maps allocated in GPU memory for use during back-propagation.
The gradients of convolutional features (and parameters) typically are functions of the intermediate outputs, and therefore the intermediate outputs must remain accessible for the backward pass.
In DenseNets, there are two operations which are responsible for this quadratic growth: pre-activation batch normalization and contiguous concatenation.


\begin{figure}
  \begin{subfigure}[b]{.48\linewidth}
  \centerline{
    \includegraphics[width=0.8\linewidth]{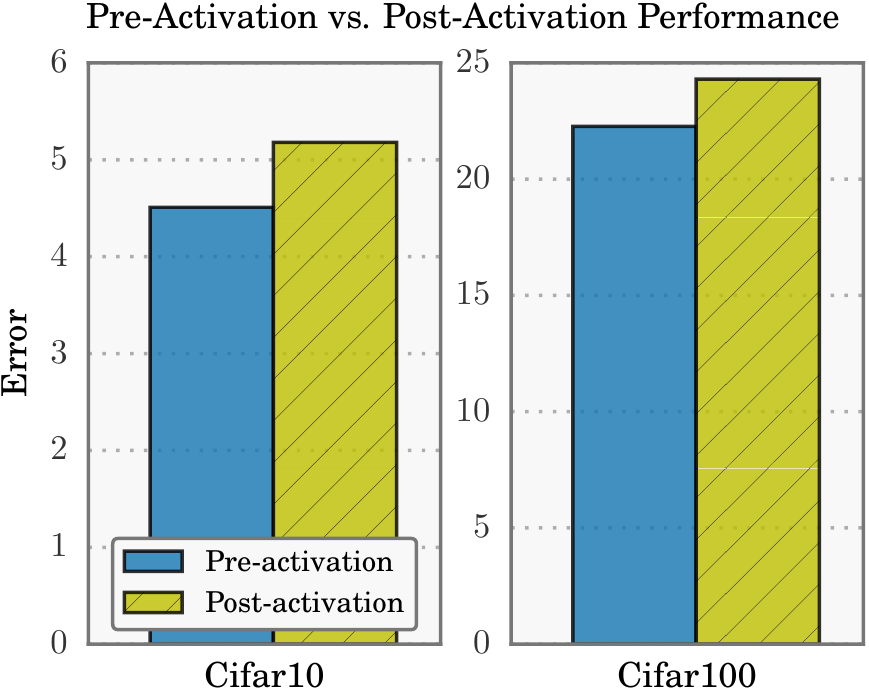}}
  \end{subfigure}
  \quad
  \begin{subfigure}[b]{.48\linewidth}
    \centerline{\includegraphics[width=0.8\linewidth]{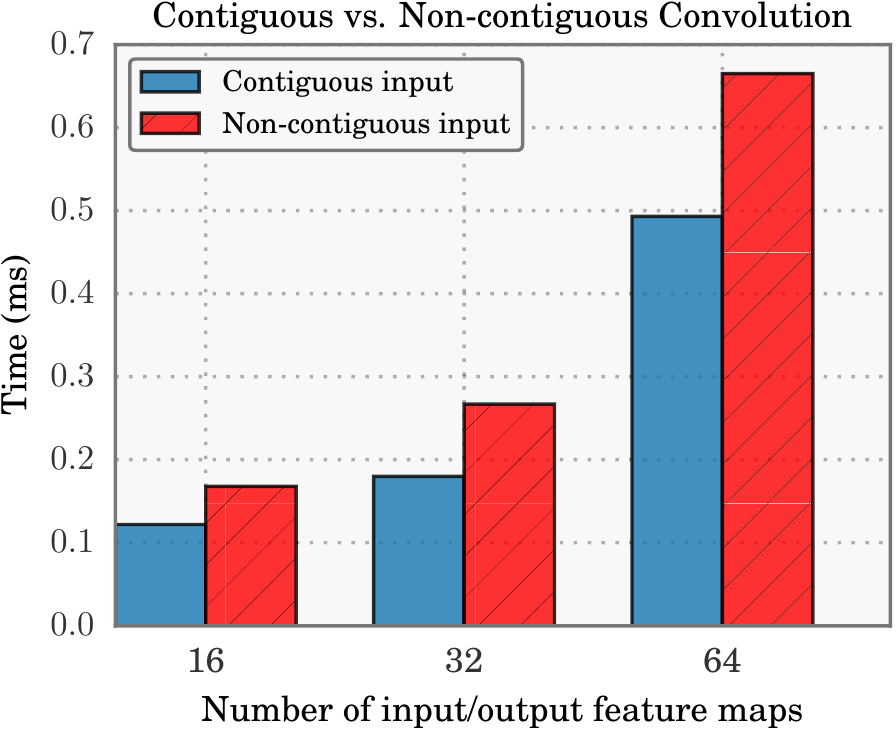}}
  \end{subfigure}
  \caption{
    {\bf Left:} Comparison of pre-activation vs post-activation DenseNet architectures (100-layer DenseNet-BC, $k=12$).
      Pre-activations offer significant error reductions, but generate a quadratic number of outputs.
    {\bf Right:} Speed of contiguous vs non-contiguous convolution operations (measured on a NVIDIA GTX 980).
    Each operation applies $3$-by-$3$ filters to $32 \times 32$ feature maps (minibatch size of $64$).
    Contiguous operations are preferred, but generate multiple copies of each feature.
  }
  \label{fig:performance_of_architectures}
\end{figure}


\paragraph{Pre-activation batch normalization.}
The DenseNet architecture, as described in \cite{huang2017densely}, utilizes \emph{pre-activation} batch normalization \cite{he2016identity}.
Unlike conventional architectures, pre-activation networks apply batch normalization and non-linearities \emph{before} the convolution operation rather than after.
Though this might seem like a minor change, it makes a big difference in DenseNet performance.
Batch normalization applies a scaling and a bias to the input features.
If each layer applies its own batch normalization operation, then each layer applies a \emph{unique} scale and bias to previous features.
For example, the Layer 2 batch normalization might scale a Layer 1 feature by a positive constant, while Layer 3 might scale the same feature by a negative constant.
After the non-linearity, Layer 2 and Layer 3 extract opposite information from Layer 1.
This would not be possible if all layers shared the same batch normalization operation, or if normalization occurred after convolution operations.
Without pre-activation, the CIFAR-10 error of a 100-layer DenseNet-BC grows from $4.51$ to $5.18$.
On CIFAR-100, the error grows from $22.27$ to $24.30$ (\autoref{fig:performance_of_architectures} left).

Given a DenseNet with $m$ layers, pre-activations generate up to $m$ normalized copies of each layer.
Because each copy has different scaling and bias, na\"ive implementations in standard deep learning frameworks typically allocate memory for each of these $(m-1)(m-2)/2$ duplicated feature maps. 



\paragraph{Contiguous concatenation.}
Convolution operations are most efficient when all input data lies in a \emph{contiguous block of memory}.
Some deep learning frameworks, such as Torch, explicitly require that all convolution operations are contiguous.
While CUDNN, the most common library for low-level convolution routines \cite{cudnn}, does not have this requirement, using non-contiguous blocks of memory adds a computation time overhead of $30-50\%$ (\autoref{fig:performance_of_architectures} right).

To make a contiguous input, each layer must copy all previous features into a contiguous memory block.
Given a network with $m$ layers, each feature may be copied up to $m$ times. If these copies are stored separately, the concatenation operation would also incur quadratic memory cost.

It is worth noting that we cannot simply assign filter outputs to a pre-allocated contiguous block of memory.
Features are represented as tensors in $\mathbb{R}^{n \times d \times w \times h}$ (or $\mathbb{R}^{n \times w \times h \times d}$), where $n$ is the number of mini-batch samples, $d$ is the number of feature maps, and $w$ and $h$ are the width and height.
GPU convolution routines, such as from the CUDNN library, assume that feature data is stored with the minibatch as the outer dimension.
Assigning two features next to each other in a contiguous block therefore concatenates along the minibatch dimension, and not the intended feature map dimension.

\section{Na\"ive Implementation}
\label{sec:naive_implementation}

\begin{figure}
  \centering
  \includegraphics[width=\linewidth]{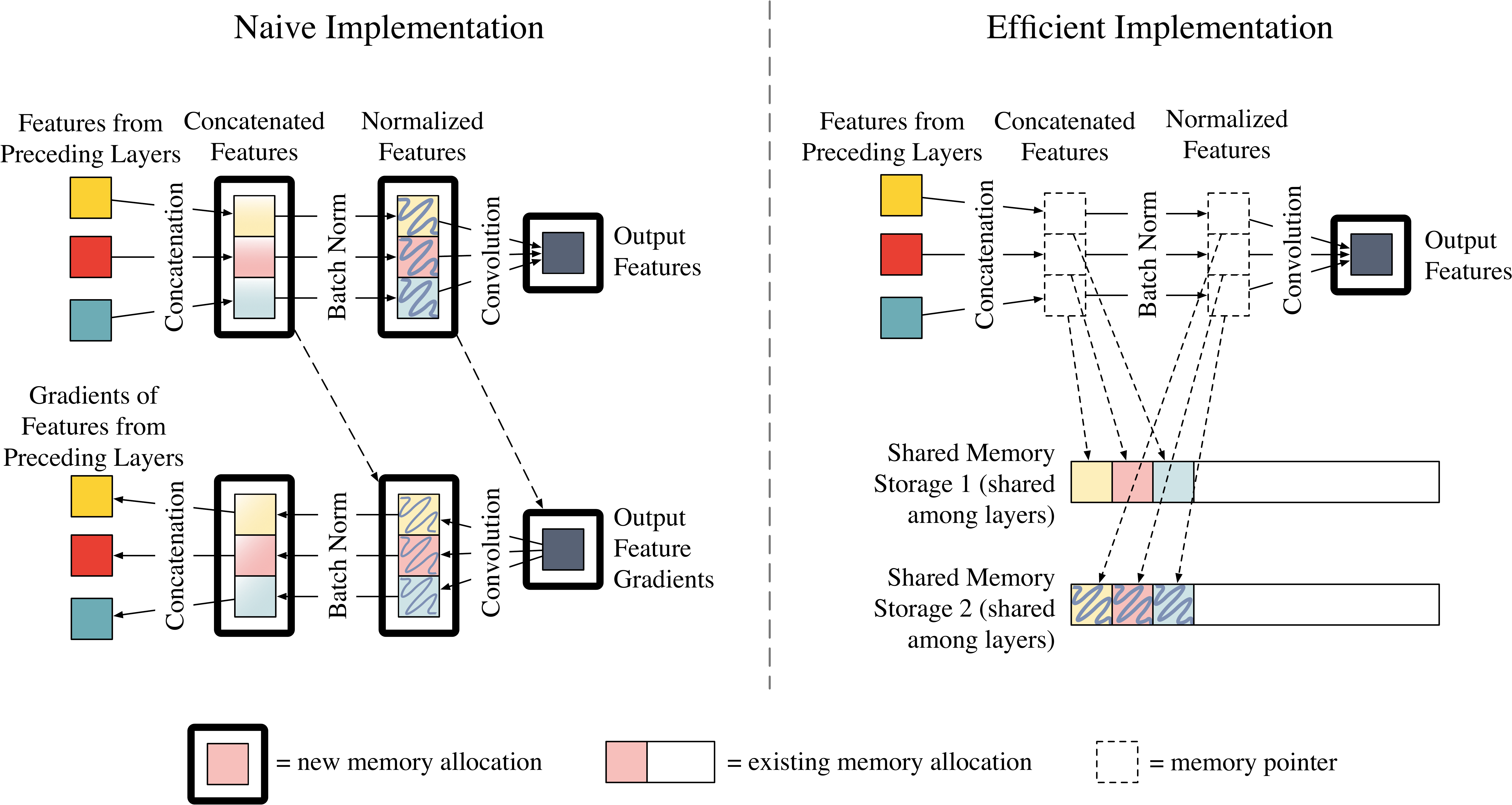}
  \caption{
    DenseNet layer forward pass: original implementation ({\bf left}) and efficient implementation ({\bf right}).
    Solid boxes correspond to tensors allocated in memory, where as translucent boxes are pointers.
    Solid arrows represent computation, and dotted arrows represent memory pointers.
    The efficient implementation stores the output of the concatenation, batch normalization, and ReLU layers in temporary storage buffers, whereas the original implementation allocates new memory.
  }
  \label{fig:forward}
\end{figure}

In modern deep learning libraries, layer  operations are typically represented by edges in a computation graph.
Computation graph nodes represent intermediate feature maps stored in memory.
We show the computation graph for a na\"ive implementation of a DenseNet layer (without bottleneck operations) in \autoref{fig:forward} (top left).\footnote{
  The computation graph is based roughly on the original implementation:
  \url{https://github.com/liuzhuang13/DenseNets/}.
}
The inputs to each layer are features from previous layers (the colored boxes).
As these features originate from different layers, they are not stored contiguously in memory. The first operation of the na\"ive implementation therefore \emph{copies} each of these features and concatenates them into a contiguous block of memory (center left).
If each of the $\ell$ previous layers produces $k$ features, the contiguous block of memory must accommodate $\ell \times k$ feature maps.
These concatenated features are input to the batch normalization operation (center right), which similarly allocates a new $\ell \times k$ contiguous block of memory.
(The ReLU non-linearity occurs in-place in memory, and therefore we choose to exclude it from the computation graph for simplicity.)
Finally, a convolution operation (far right) generates $k$ new features from the batch normalization output.
From \autoref{fig:forward} (top left), it is easy to visualize the quadratic growth of memory.
The two intermediate operations require a memory block which can fit all $O(\ell k)$ previously computed features.
By comparison, the output features require only a constant $O(k)$ memory per layer.

In some deep learning frameworks, such as LuaTorch, even more memory may be allocated during back-propagation.
\autoref{fig:forward} (bottom left) displays a computation graph for gradients.
The output feature gradients (far right) and the normalized feature maps from the forward pass (dotted line) are used to compute the batch normalization gradients (center right).
Storing these gradients requires an additional $\ell \times k$ feature maps worth of memory allocation.
Similarly, a $\ell \times k$ memory allocation is necessary to store the concatenated feature gradients.

\section{Memory-Efficient Implementation}
\label{sec:mem-efficient}

\newcommand{\bigo}[1]{O \! \left( #1 \right)}

To circumvent this issue we exploit that concatenation and normalization operations are computationally extremely cheap. We propose two  pre-allocated \emph{Shared Memory Storage} locations to avoid the quadratic memory growth.  
During the forward pass, we assign all intermediate outputs to these memory blocks.  During back-propagation, we recompute the concatenated and normalized features on-the-fly as needed. 

This recomputation strategy has previously been explored on other neural network architectures.
\citet{chen2016training} exploit recomputation to train $1,000$-layer ResNets on a single GPU.
Additionally, \citet{mxnet} have developed recomputation support for arbitrary networks in the MxNet deep learning framework.
In general, recomputing intermediate outputs necessarily trades-off memory for computation. However, we have discovered that this  strategy is very practical for DenseNets.
The concatenation and batch normalization operations are responsible for most memory usage, yet only incur a small fraction of the overall computation time. 
Therefore, recomputing these intermediate outputs yields substantial memory savings with little computation time overhead.

\paragraph{Shared storage for concatenation.}
The first operation -- feature concatenation -- requires $\bigo{\ell k}$ memory storage, where $\ell$ is the number of previous layers and $k$ is the number of features added each layer.
Rather than allocating memory for each concatenation operation, we assign the outputs to a memory allocation shared across all layers (``Shared Memory Storage 1'' in \autoref{fig:forward} right).
The concatenation feature maps (center left) are therefore pointers to this shared memory.
Copying to pre-allocated memory is significantly faster than allocating new memory, so this concatenation operation is extremely efficient.
The batch normalization operation, which takes these concatenated features as input, reads directly from Shared Memory Storage 1.
Because Shared Memory Storage 1 is used by all network layers, its data is not permanent.
When the concatenated features are needed during back-propagation, we assume that they can be recomputed efficiently.

\paragraph{Shared storage for batch normalization.}
Similarly, we assign the outputs of batch normalization (which also requires $\bigo{m}$ memory) to a shared memory allocation (``Shared Memory Storage 2'').
The convolution operation reads from pointers to this shared storage (center right). As with concatenation, we must recompute the batch normalization outputs during back-propagation, as the data in Shared Memory Storage 2 is not permanent and will be overwritten by the next layer.
However, batch normalization consists of scalar multiplication and addition operations, which are very efficient in comparison with convolution math.
Computing Batch Normalized features accounts for roughly $5\%$ of one forward-backward pass, and therefore is not a costly operation to repeat.

\paragraph{Shared storage for gradients.}
The concatenation, batch normalization, and convolution operations each produce gradients tensors during back-propagation.
In LuaTorch, it is straight forward to ensure that this tensor data is stored in a single shared memory allocation.
This strategy prevents gradient storage from growing quadratically.
We use a memory-sharing scheme based on of Facebook's ResNet implementation.\footnote{
  \url{https://github.com/facebook/fb.resnet.torch}
}
The PyTorch and MxNet frameworks share gradient storage out-of-the-box.

\paragraph{Putting these pieces together,} the forward pass (\autoref{fig:forward} right) is similar to the na\"ive implementation (\autoref{fig:forward} top left), with the exception that intermediate feature maps are stored in Shared Memory Storage 1 or Shared Memory Storage 2.
The backward pass requires one additional step: we first re-compute the concatenation and batch normalization operations (center left and center right) in order to re-populate the shared memory storage with the appropriate feature map data.
Once the shared memory storage contains the correct data, we can perform regular back-propagation to compute gradients.
In total, this DenseNet implementation only allocates memory for the output features (far right), which are constant in size. Its overall memory consumption for feature maps is \emph{linear} in network depth.

\section{Results}

\begin{figure}
  \centering
  \includegraphics[width=\linewidth]{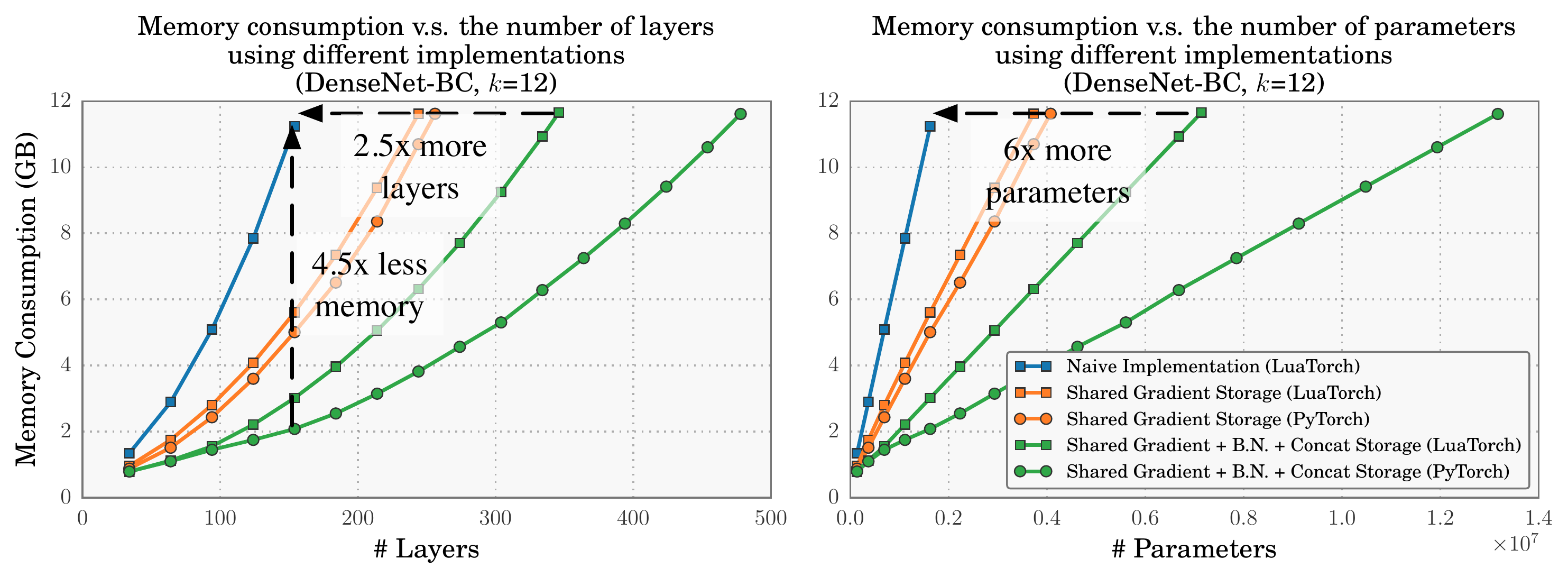}
  \caption{
    GPU memory consumption as a function of network depth.
    Each model is a bottlenecked Densenet (Densenet-BC) with $k=12$ features added per layer.
    The efficient implementations enables training much deeper models with less memory.
  }
  \label{fig:memvslayer}
\end{figure}

We compare the memory consumption and computation time of three DenseNet implementations during training.
The {\bf na\"ive implementation} is based on the original LuaTorch implementation\footnote{
  \url{https://github.com/liuzhuang13/DenseNet}
} of \citet{huang2017densely}.
As described in \autoref{sec:naive_implementation}, this implementation allocates memory for the concatenation and batch normalization outputs.
Additionally, each of these operations uses additional memory to store gradients of the intermediate features.
Therefore, this implementation has four operations with quadratic memory growth (two forward-pass operations and two backward-pass operations).

We then test two memory-efficient implementations of DenseNets.
The first implementation {\bf shares gradient storage} so that there are no new memory allocations during back-propagation. All gradients are instead assigned to shared memory storage.
The LuaTorch gradient memory-sharing code is based on Facebook's ResNet implementation\footnote{
  \url{https://github.com/facebook/fb.resnet.torch}
}.
PyTorch automatically performs this optimization out-of-the-box.
The second implementation includes all optimizations described in \autoref{sec:mem-efficient}: shared storage for batch normalization, concatenation, and gradient operations ({\bf shared gradient + B.N. + concat storage}).
The concatenated and normalized feature maps are recomputed as necessary during back-propagation.
The only tensors which are stored in memory during training are the convolution feature maps and the parameters of the network.

\paragraph{Memory consumption} of the three implementations (in LuaTorch and PyTorch) is shown in \autoref{fig:memvslayer}.
(There is no PyTorch na\"ive implementation because PyTorch automatically shares gradient storage.)
With four quadratic operations, the na\"ive implementation becomes memory intensive very quickly.
The memory usage of a $160$ layer network ($k=12$ features per layer, $1.8M$ parameters) is roughly $10$ times as much as a $40$ layer network ($k=12$, $160K$ parameters).
Training a larger network with more than $160$ layers requires over 12 GB of memory, which is more than a typical single GPU.
While sharing gradients reduces some of this memory cost, memory consumption still grows rapidly with depth.
On the other hand, using all the memory-sharing operations significantly reduces memory consumption. In LuaTorch, the $160$-layer model uses $22\%$ of the memory required by the Na\"ive Implementation. Under the same memory budget (12 GB), it is possible to train a $340$-layer model, which is $2.5\times$  as deep and has $6\times$ as many parameters as the best na\"ive implementation model.

\begin{wrapfigure}{r}{0.5\textwidth}
  \centering
  \includegraphics[width=\linewidth]{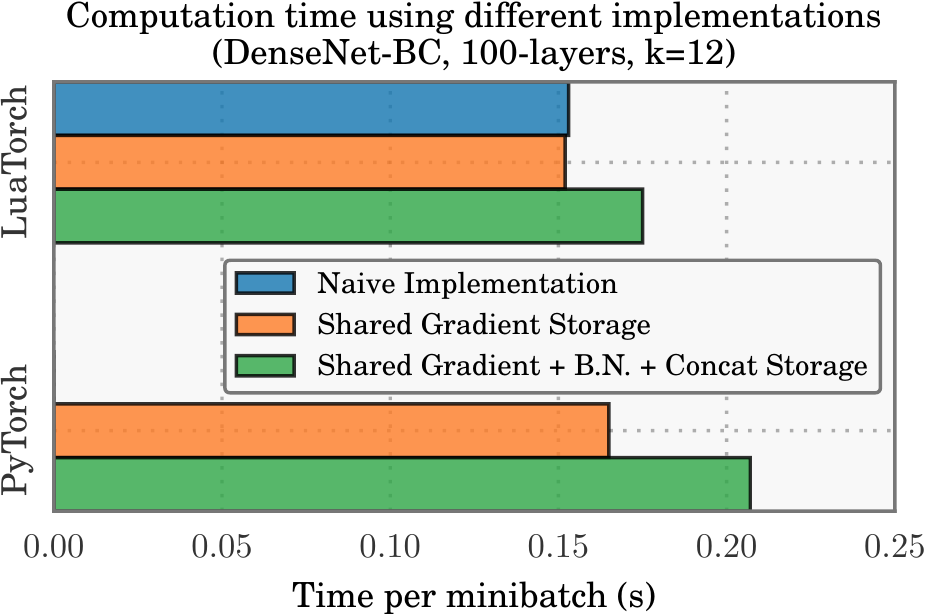}
  \caption{
    Computation time (measured on a NVIDIA Maxwell Titan-X).
  }
  \label{fig:timevslayer}
\end{wrapfigure}

It is worth noting that the \emph{total} memory consumption of the most efficient implementation does not grow linearly with depth, as the number of parameters is inherently a quadratic function of the network depth. This is a function of the architectural design and part of the reason why DenseNets are so efficient. The memory required to store the parameters is far less than the memory consumed by the feature maps and the remaining quadratic term  does not impede model depth.


Finally, we see in \autoref{fig:memvslayer} that PyTorch is more memory efficient than LuaTorch.
Using the efficient PyTorch implementation, we can train DenseNets with nearly $500$ layers ($13M$ parameters) on a single GPU.
The ``autograd'' library in PyTorch performs memory optimizations during training, which likely contribute to this implementation's efficiency.

\paragraph{Training time} is not significantly affected by the memory optimizations.
In \autoref{fig:timevslayer} we plot the time per minibatch of a 100-layer DenseNet-BC ($k=12$) on a NVIDIA Maxwell Titan-X.
Shared gradient storage does not incur any time cost.
Sharing batch normalization and concatenation storage adds roughly $15\%$ time overhead on LuaTorch, and $20\%$ on PyTorch.
This extra cost is a result of recomputing operations during back-propagation.
For DenseNets of any size, it makes sense to share gradient storage.
If GPU memory is limited, the time overhead from sharing batch normalization/concatenation storage constitutes a reasonable trade-off.

\begin{figure}
  \includegraphics[width=\linewidth]{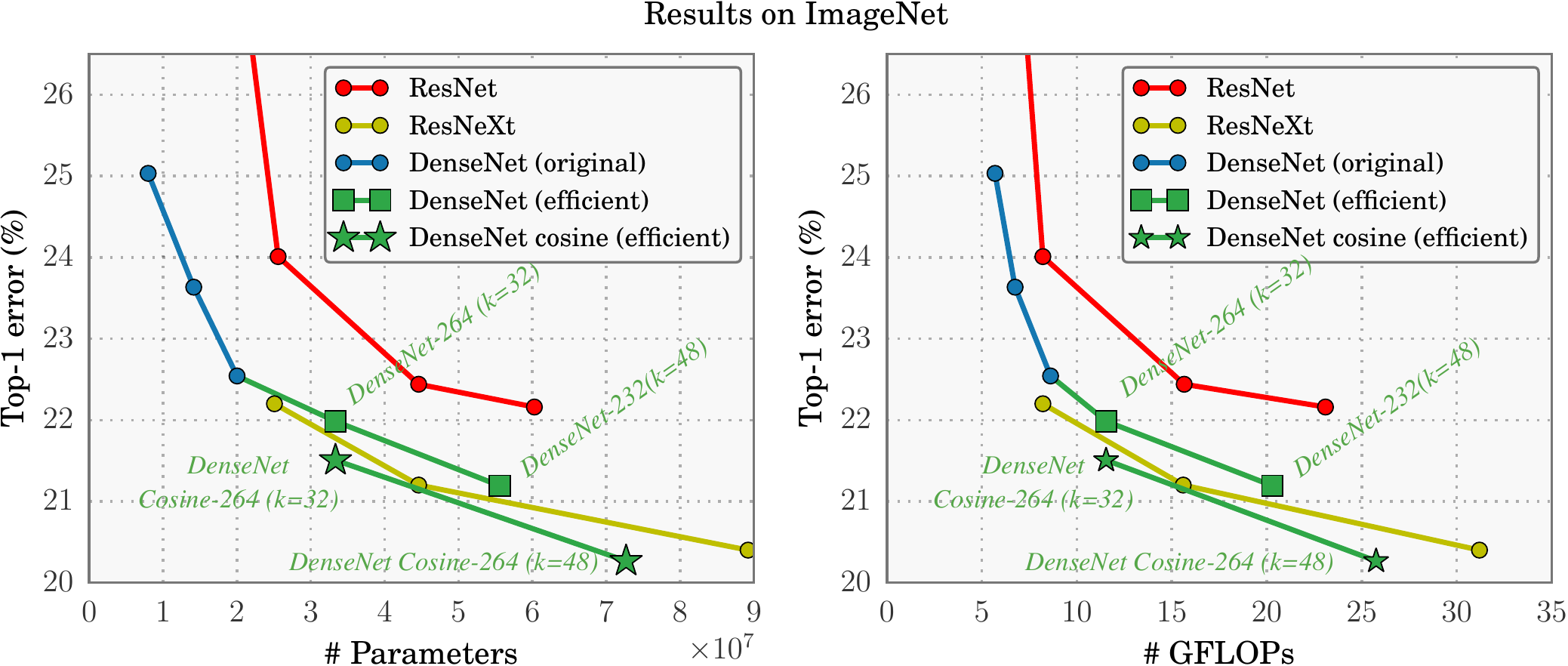}
  \caption{
    Top-1 classification error on ImageNet.
    The DenseNet models were trained on 8 NVIDIA Tesla M40 GPUs.
    Results in stars were not possible to train without the efficient implementation.
  }
  \label{fig:results}
\end{figure}

\paragraph{ImageNet results.}
We test the new memory efficient LuaTorch implementation (shared memory storage for gradients, batch normalization, and concatenation) on the ImageNet IVLSRC classification dataset \cite{russakovsky2015imagenet}.
The deepest model trained by \citet{huang2017densely} using the original LuaTorch implementation was $161$ layers ($k=48$ features per layer, $29M$ parameters).
With the efficient LuaTorch implementation however, it is possible to train a $264$-layer DenseNet ($k=48$, $73M$ parameters) using 8 NVIDIA Tesla M40 GPUs.

We train two new {\bf DenseNet} models with the efficient implementation, one with $264$ layers ($k=32$, $33M$ parameters) and one with $232$ layers ($k=48$, $55M$ parameters).\footnote{
  There are four dense blocks in this model.
  In the $264$-layer model, the dense blocks have 6, 32, 64, and 48 layers respectively.
  In the $232$-layer model, the dense blocks have 6, 32, 48, and 48 layers.
}
These models are trained for 100 epochs following the same procedure described in \cite{huang2017densely}.
Additionally, we train two DenseNets models with a cosine learning-rate schedule ({\bf DenseNet cosine}), similar to what was used by \cite{loshchilov2016sgdr} and \cite{huang2017snapshot}.
The model is trained for 100 epochs, with the learning rate of epoch $t$ set to $0.05 \cos(t (\pi / 100)) + 1$.
Intuitively, this learning rate schedule starts off with a large learning rate, and quickly (but smoothly) anneals the learning rate to a small value.
Both of the cosine DenseNets have $264$ layers -- one with $k=32$ ($33M$ parameters) and one with $k=48$ ($73M$ parameters).
Given a fixed GPU budget, these DenseNet models can only be trained with the efficient implementation.

We display the top-1 error performance of these DenseNets in \autoref{fig:results}.
The models trained with the efficient implementation are denoted as green points (squares for standard learning rate schedule, stars for cosine).
We compare the performance of these models to shallower DenseNets trained with the original implementation.
Additionally, we compare against {\bf ResNet} models introduced in \cite{he2016deep} and {\bf ResNeXt} models introduced in \cite{xie2016aggregated}.
Results were obtained with single centered test-image crops.

The new DenseNet models (standard training procedure) achieve nearly $1$ percentage point better error than the deepest ResNet model, while using fewer parameters.
Additionally, the deepest cosine DenseNet achieves a top-1 error of $20.26\%$, which outperforms the previous state-of-the-art model \cite{xie2016aggregated}.
It is clear from these results that DenseNet performance continues to improve measurably as more layers are added.

\section{Conclusion}

In this report, we describe a new implementation strategy for DenseNets.
Previous DenseNet implementations store all intermediate feature maps during training, causing feature map memory usage to grow quadratically with depth.
By employing a shared memory buffer and recomputing some cheap transformations, models utilize significantly less memory, with only a small increase in computation time.
With this new implementation strategy, memory no longer impedes training extremely deep DenseNets.
As a result, we are able to double the depth of prior models, which results in a measurable drop in top-1 error on the ImageNet dataset.

\section*{Acknowledgments}
The authors are supported in part by the III-1618134, III-1526012, and IIS-1149882 grants from the
National Science Foundation, 
the Office of Naval Research DOD (N00014-17-1-2175),
and the Bill and Melinda Gates Foundation.

{\footnotesize
  \bibliographystyle{abbrvnat}
  \bibliography{citations}
}

\end{document}